\documentclass[11pt,a4paper]{article}
\usepackage[hyperref]{emnlp-ijcnlp-2019}
\usepackage{times}
\usepackage{latexsym}

\usepackage{graphicx}
\usepackage{subcaption}
\usepackage{booktabs}

\usepackage{url}
\usepackage{color}

\aclfinalcopy 

\title{Hard but Robust, Easy but Sensitive: 
	How Encoder and Decoder Perform in Neural Machine Translation}

\author{Tianyu He \thanks{\ \ \ The work was conducted at Microsoft Research Asia.} \ $^1$ ,\ Xu Tan $^2$ \and Tao Qin $^2$ \\
  $^1$University of Science and Technology of China \\
  $^2$Microsoft Research \\
  {\tt hetianyu@mail.ustc.edu.cn},\quad
  {\tt \{xuta, taoqin\}@microsoft.com} \\}

\date{}

\begin{document}
\maketitle
\begin{abstract}
  Neural machine translation (NMT) typically adopts the encoder-decoder framework. A good understanding of the characteristics and functionalities of the encoder and decoder can help to explain the pros and cons of the framework, and design better models for NMT. In this work, we conduct an empirical study on the encoder and the decoder in NMT, taking Transformer as an example. We find that 1) the decoder handles an easier task than the encoder in NMT, 2) the decoder is more sensitive to the input noise than the encoder, and 3) the preceding words/tokens in the decoder provide strong conditional information, which accounts for the two observations above. We hope those observations can shed light on the characteristics of the encoder and decoder and inspire future research on NMT.
\end{abstract}

\section{Introduction}
\label{sec:intro}
The encoder-decoder based framework~\citep{sutskever2014sequence,bahdanau2014neural,luong2015effective} is the dominant approach for neural machine translation (NMT)~\citep{vaswani2017attention, hassan2018achieving}. Although the encoder and decoder usually adopt the same model structure (RNN~\citep{wu2016google}, CNN~\citep{gehring2017convolutional} or self-attention~\citep{vaswani2017attention,he2018layer}) and the same number of layers, they perform different functionalities: the encoder extracts the hidden representations of the source sentence, and the decoder generates target tokens conditioned on the source hidden representations as well as the previous generated tokens.

While most existing works focus on the design and improvement of encoder-decoder framework for NMT~\citep{kaiser2018depthwise,gehring2017convolutional,vaswani2017attention,dehghani2018universal} as well as its detailed analyses~\citep{ding2017visualizing,ghader2017does,domhan2018much,tang2018analysis},  
few works concentrate on the characteristics and functionalities of the encoder and the decoder, which are valuable to understand this popular framework and improve its performance in NMT. Therefore, in this paper, we conduct a study and aim to understand the characteristics of the encoder and the decoder in NMT. We observe some interesting phenomena:
\begin{itemize}
	\item The decoder handles an easier task than the encoder. 1) We find that adding more layers to the encoder achieves larger improvements than adding more layers to the decoder. 2) We also compare the training time of the encoder and decoder by fixing the parameters of a well-trained decoder (encoder), and just update the parameters of the encoder (decoder). We found that the decoder converges faster than the encoder. These two results suggest that the decoder handles an easier task than the encoder in NMT.
	\item The decoder is more sensitive to the input noise than the encoder. We randomly add different level of noise to the input of the encoder and decoder respectively during inference, and find that adding noise to the input of the decoder leads to better accuracy drop than that of the encoder. 
	\item We further analyze why the decoder is more sensitive by masking the previous tokens, and comparing autoregressive NMT with the non-autoregressive counterpart. We find that the preceding tokens in the decoder provide strong conditional information, which partially explain the previous two  observations on the decoder.
\end{itemize}

We believe our studies on the different characteristics of the encoder and decoder will inspire the following research on the encoder-decoder framework as well as improve the performance on NMT and other encoder-decoder based tasks.

\section{Related Work}
\label{sec:related}

In this section, we mainly review the work on the analysis of encoder-decoder framework in the field of NMT. \citet{ding2017visualizing} analyzed how NMT works based on the encoder-decoder framework and explained the translation errors with the proposed layer-wise relevance propagation. \citet{belinkov2017evaluating} focused on the representation of different layers of the encoder in NMT. \citet{belinkov2017synthetic} concluded that NMT systems are brittle.
In the view of attention mechanism, \citet{ghader2017does} studied the encoder-decoder attention and provided detailed analysis of what is being learned by the attention mechanism in NMT model. \citet{song2018hybrid} studied the effectiveness of hybrid attention in neural machine translation. \citet{shen2018dense} introduced dense connection in encoder-decoder attention. \citet{tang2018analysis} analyzed the encoder-decoder attention mechanisms in the case of word sense disambiguation. For model architecture, \citet{britz2017massive} provided early study on RNN-based architectural hyperparameters. \citet{song2019mass} proposed to pre-train the encoder-attention-decoder framework jointly. \citet{domhan2018much} conducted fine-grained analyses on various modules and made comparisons between RNN, CNN and self-attention.
While the works above provide a better understanding of NMT model from the perspective of attention mechanism, network architecture and language modeling, few of them focus on the comparison of the characteristics for the encoder and the decoder in NMT. This is exactly what we focus in this paper.

\section{Hard for Encoder, Easy for Decoder}
\label{sec:hard_easy}

In general, the encoder and decoder perform different functionalities in an NMT model. In this section, we compare the characteristics between the encoder and decoder by analyzing the difficulty of the corresponding task they handle in NMT. We investigate the task difficulty by comparing the training effort of the encoder and decoder from two perspectives: the number of layers and the convergence speed for the encoder and decoder respectively.

We train the Transformer~\citep{vaswani2017attention}\footnote{We choose Transformer as the basic model structure since it achieves the state-of-the-art performance and becomes the most popular structure for recent NMT research.} on IWSLT14 German$\leftrightarrow$English (De$\leftrightarrow$En) and Romanian$\leftrightarrow$English (Ro$\leftrightarrow$En) translation.
The translation quality is measured by BLEU score. More details on datasets, model configurations and evaluation metrics can be found in the supplementary materials (Section 1.1).

\subsection{Deeper Encoder Brings More Gains Than Deeper Decoder}

\begin{figure*}[ht]
	\small
	\centering
	\begin{subfigure}[t]{0.329\textwidth}
		\centering
		\includegraphics[width=\textwidth]{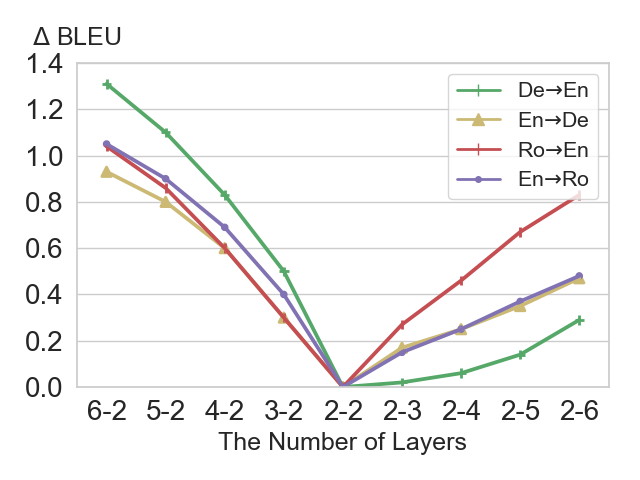}
		\caption{\small Increase the number of layers.}
		\label{fig:diff_layers_increa}
	\end{subfigure}
	\begin{subfigure}[t]{0.329\textwidth}
		\centering
		\includegraphics[width=\textwidth]{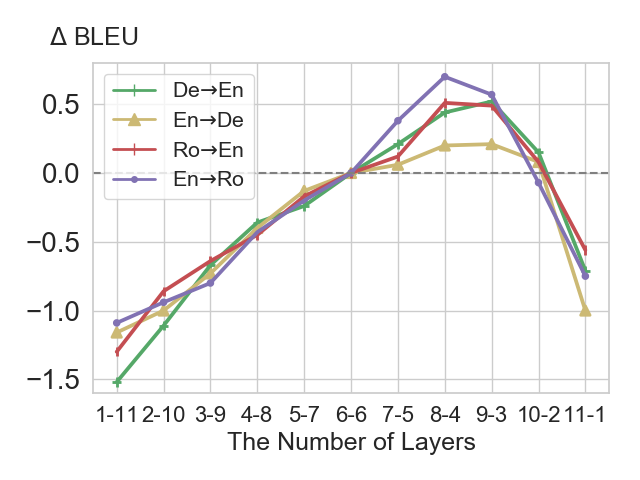}
		\caption{\small The total number of layers is fixed.}
		\label{fig:diff_layers_fix}
	\end{subfigure}
	\begin{subfigure}[t]{0.329\textwidth}
		\centering
		\includegraphics[width=\textwidth]{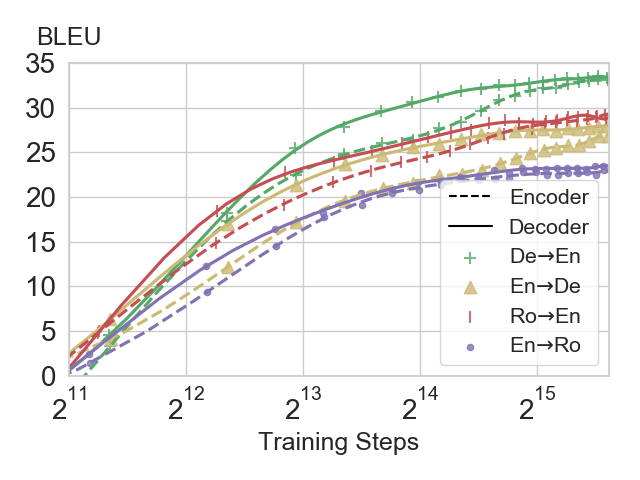}
		\caption{\small The convergence speed.}
		\label{fig:train_time}
	\end{subfigure}
	\caption{\small Experimental results for analyzing the training effort for the encoder and decoder. X-axes in Figure~\ref{fig:diff_layers_increa} and~\ref{fig:diff_layers_fix} indicate the number of layers for encoder and decoder respectively (Encoder-Decoder). Y-axes in Figure~\ref{fig:diff_layers_increa} and~\ref{fig:diff_layers_fix} indicate the delta BLEU to the 2-2 layer model (Figure~\ref{fig:diff_layers_increa}) and the 6-6 layer model (Figure~\ref{fig:diff_layers_fix}).}
	\label{fig:hard_easy}
\end{figure*}

\begin{figure*}[ht]
	\small
	\centering
	\begin{subfigure}[t]{0.329\textwidth}
		\centering
		\includegraphics[width=\textwidth]{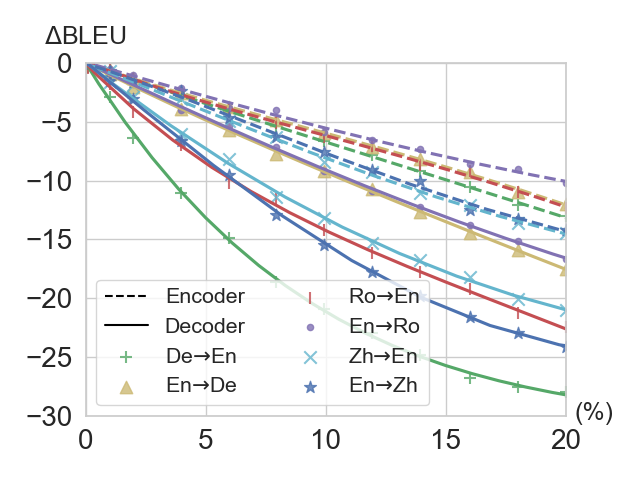}
		\caption{\small Random dropping}
	\end{subfigure}
	\begin{subfigure}[t]{0.329\textwidth}
		\centering
		\includegraphics[width=\textwidth]{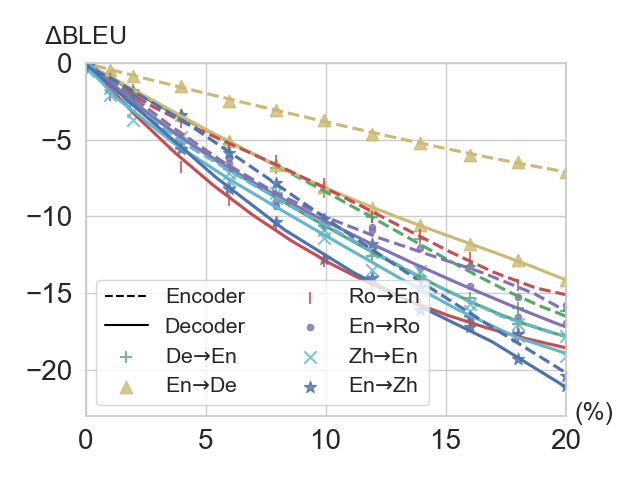}
		\caption{\small Random noising}
	\end{subfigure}
	\begin{subfigure}[t]{0.329\textwidth}
		\centering
		\includegraphics[width=\textwidth]{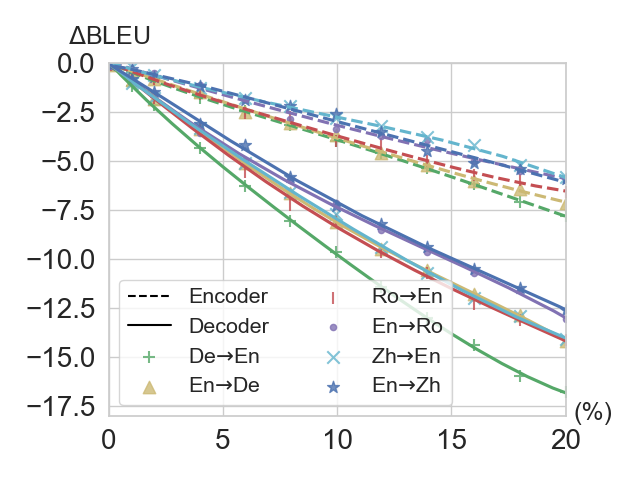}
		\caption{\small Random swapping}
	\end{subfigure}
	\caption{\small The accuracy drop (compared to the model with no input noise) when adding noise to the input tokens of the encoder and decoder. X-axis indicates the perturbation rate (\%), i.e., the ratio of tokens being perturbed by noise.}
	\label{fig:random_disturb}
\end{figure*}

\begin{figure*}[t]
	\small
	\centering
	\begin{subfigure}[t]{0.329\textwidth}
		\centering
		\includegraphics[width=\textwidth]{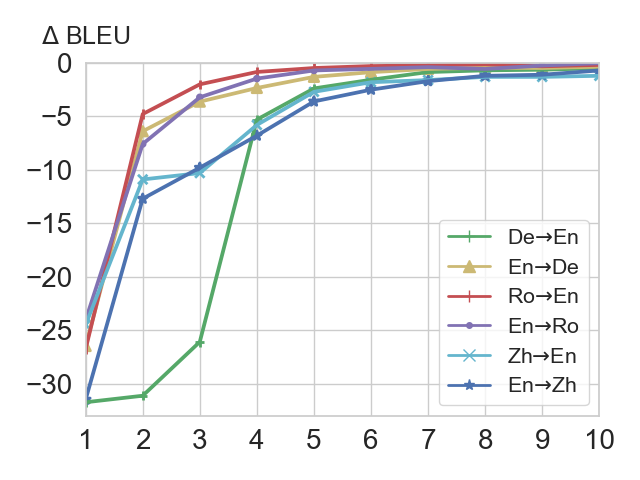}
		\caption{\small The translation quality sharply declines when dropping nearby tokens.}
		\label{fig:distance}
	\end{subfigure}
	\begin{subfigure}[t]{0.329\textwidth}
		\centering
		\includegraphics[width=\textwidth]{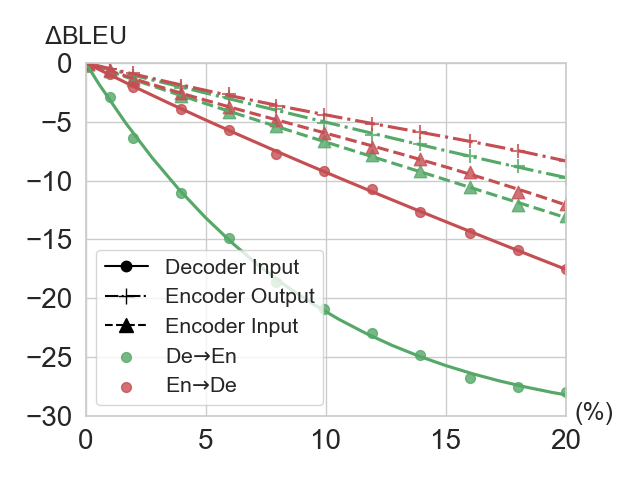}
		\caption{\small Randomly drop tokens from the three places for autoregressive NMT.}
		\label{fig:at}
	\end{subfigure}
	\begin{subfigure}[t]{0.329\textwidth}
		\centering
		\includegraphics[width=\textwidth]{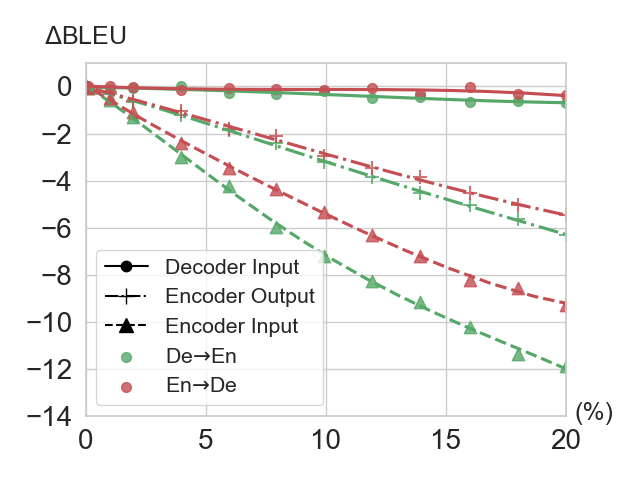}
		\caption{\small Randomly drop tokens from the three places for non-autoregressive NMT.}
		\label{fig:nat}
	\end{subfigure}
	\caption{\small Further analyses on characteristics of the encoder and decoder. X-axis in Figure~\ref{fig:distance} indicates the distance between the dropped token and the current predicted token. X-axes in Figure~\ref{fig:at} and \ref{fig:nat} indicate the perturbation rate (\%). See text for the detailed description.}
	\label{fig:further_anal}
\end{figure*}

We vary the number of layers for the encoder and decoder respectively, to investigate which can achieve more gains with deeper layers. Deeper encoder (decoder) with more gains can indicate the task that encoder (decoder) handles needs more representation capability, which is harder to learn. 

We consider two different strategies when varying the number of layers: 1) increasing the number of layers for the encoder and decoder respectively based on a baseline model with a $2$-layer encoder and a $2$-layer decoder; 2) adjusting the number of layers for the encoder and decoder under the constraint that the total number of layers for encoder and decoder is fixed (totally $12$ layers in our experiments).

The results are demonstrated in Figure~\ref{fig:diff_layers_increa} and~\ref{fig:diff_layers_fix}. We have two observations: 1) adding more layers to the encoder brings more gain than the decoder (Figure~\ref{fig:diff_layers_increa}); 2) the model with a shallow encoder typically performs worse than the model with a shallow decoder (Figure~\ref{fig:diff_layers_fix}).


\subsection{Encoder Converges Slower than Decoder}
We further investigate the difficulty of the task for the encoder and decoder by comparing their convergence speed. Encoder (decoder) that needs more training time indicates the task that encoder (decoder) handles is more difficult.

We evaluate the convergence speed for the encoder by initializing the model with the parameters of a well-trained decoder and fixing them during training, and initializing other components with random variables. We follow the opposite practice when evaluating the convergence speed for the decoder. We show the BLEU scores along the training process on all the four datasets in Figure~\ref{fig:train_time}, which demonstrates that the decoder converges faster than the encoder.

\paragraph{Discussion} From the analyses in the last two subsections, it can be seen that increasing the layer of encoder brings more gains than increasing the layer of the decoder, and encoder needs more training time than the decoder, which demonstrate that the encoder of the NMT model handles a more difficult task than the decoder.

\section{Robust Encoder, Sensitive Decoder}
\label{sec:robust_sensitive}

In this section, we compare the characteristics between the encoder and decoder by analyzing  their robustness according to the input noise in the inference phase. We simulate the input noise with three typical operations~\citep{artetxe2018unsupervised,lample2018unsupervised}: 1) random dropping: we randomly drop the input tokens of encoder and decoder respectively with different drop rates; 2) random noising: we randomly select tokens and replace its embedding with random noise; 3) random swapping: we randomly reverse the order for the adjacent tokens. 
The decoder in NMT model typically generates the current token one-by-one conditioned on the previous generated tokens, which suffers from error propagation~\citep{bengio2015scheduled}: if a token is incorrectly predicted by the decoder, it will affect the prediction of the following tokens. Adding input noise to the decoder will further enhance the effect of error propagation, and thus influence our analysis. To eliminate the influence of error propagation, we apply teacher forcing~\cite{williams1989learning} in the inference phase by feeding the previous ground-truth target tokens instead of the previously generated target tokens, following ~\citet{wu2018beyond}. 

We evaluate our model on IWSLT14 De$\leftrightarrow$En, IWSLT14 Ro$\leftrightarrow$En and WMT17 Chinese$\leftrightarrow$English (Zh$\leftrightarrow$En) translation tasks. More details on experimental configurations are described in supplementary materials (Section 1.2). The results are demonstrated in Figure~\ref{fig:random_disturb}. It can be seen that as the perturbation rate increases, adding different types of noise to the decoder input consistently achieves lower translation quality than adding noise to the encoder input.

\paragraph{Discussion}
The consistent observations above suggest that the decoder is much more sensitive to the input noise than the encoder. Intuitively, the encoder aims at extracting abstract representations of the source sentence instead of depending on certain input tokens for prediction as the decoder does, demonstrating that the encoder is more robust than the decoder.

\section{Further Analysis}
Given the observations in Section~\ref{sec:hard_easy} and~\ref{sec:robust_sensitive}, we find that the decoder is more sensitive to the input tokens compared with the encoder, and the task that decoder handles is easier. In this section, we give an explanation on this phenomenon. More details on experimental configurations are described in supplementary materials (Section 1.3). Besides, we also investigate which kind of input tokens the encoder and decoder are more sensitive to, from the perspective of Part-Of-Speech (POS), and show the results in the supplementary materials (Section 2) due to space limitation.

\paragraph{Why Decoder is Easier and More Sensitive}
The decoder in NMT model typically acts as a conditional language model, which generates tokens highly depending on the previous tokens, like the standard language model~\citep{khandelwal2018sharp}. We guess the conditional information (especially the tokens right before the predicted token) is too strong for the decoder. Therefore, we study the impact of the previous tokens as follows. 

For each predicted token $w_{t}$, where $t$ is the position in the target sentence, we drop its previous token $w_{t-n}$ from the decoder input and watch the performance changes, where $n\in [1, t]$ is the distance between the dropping token and the current predicted token\footnote{We drop the same number of tokens for each $n$.}. Note that the experiments are conducted in the inference phase and evaluated with teacher forcing. As shown in Figure~\ref{fig:distance}, when dropping the token close to the predicted token, the accuracy declines more heavily than dropping the token far away, which indicates the decoder depends more on the nearby tokens.

\paragraph{Comparison with Non-Autoregressive NMT}
Since an autoregressive decoding process highly depends on previous tokens, we further investigate the characteristics of non-autoregressive decoding in NMT~\citep{gu2018nonautoregressive,guo2018non}, where each target token is generated in parallel, without dependence on previous tokens.
We compare the accuracy with different dropping rates for the encoder input, encoder output (the input of the encoder-decoder attention) and decoder input respectively, both for autoregressive NMT and non-autoregressive NMT, and illustrate the results in Figure~\ref{fig:at} and~\ref{fig:nat}. It can be observed that for autoregressive NMT, the model is more sensitive to the decoder input, and then the encoder input, and less sensitive to the encoder output. While for non-autoregressive NMT, the sensitivity is opposite to the autoregressive counterpart: the model is more sensitive to the encoder input, then encoder output, but less sensitive to the decoder input, due to the nature of non-autoregressive NMT that the decoder has to relying more on the source sentence.

\section{Conclusions and Future Work}
In this paper, we conducted a series of experiments to compare the characteristics of the encoder and decoder in NMT. We found that the decoder handles an easier task than the encoder, and the decoder is more sensitive to the input noise than the encoder. We further investigated why the decoder is more sensitive and the task it handles is easier, by analyzing the dependence of the decoder, and comparing the sensitivity to the input with non-autoregressive NMT. We hope our analyses inspire future research on NMT.

For future work, we will study the encoder-decoder framework in other machine learning tasks, such as text summarization, question answering and image captioning.


\bibliography{encdec}
\bibliographystyle{acl_natbib}


\appendix

\section{Supplemental Material}
\label{sec:supplemental}

\subsection{Detailed Experimental Settings}
In this section, we give detailed descriptions on all the experimental settings in this work.

\subsubsection{Experimental Settings for Section 3}

We train the Transformer model~\citep{vaswani2017attention}\footnote{We choose Transformer as the basic model structure since it achieves the state-of-the-art performance and becomes the most popular structure for recent NMT research.} with different number of layers on IWSLT14 German$\leftrightarrow$English (De$\leftrightarrow$En) translation task and Romanian$\leftrightarrow$English (Ro$\leftrightarrow$En) translation task.

\paragraph{Datasets}
The training sets of IWSLT14 De$\leftrightarrow$En and IWSLT14 Ro$\leftrightarrow$En contain $153K$ and $182K$ sentence pairs respectively\footnote{Data can be found at https://wit3.fbk.eu/archive/2014-01/texts}. 
We use IWSLT14.TED.tst2013 as the validation set and IWSLT14.TED.tst2014 as the test set. For De$\to$En translation, to be consistent with the previous work~\citep{edunov2018classical,ranzato2015sequence,bahdanau2016actor}, we use the same validation and test sets as those used in~\citet{edunov2018classical}, which consists of $7K$ and $7K$ sentences respectively, and follow the common practice to lowercase all words. All the datasets are preprocessed into workpieces following~\citet{wu2016google}.

\paragraph{Model Configurations}
Our experiments are based on \textit{transformer\_small} setting except for the number of layers. The model hidden size, feed-forward hidden size and the number of head are set to $256$, $1024$ and $4$ respectively. We train each model on one NVIDIA GTX1080Ti GPU until convergence and the mini-batch size is fixed as $4096$ tokens. We choose the Adam optimizer with $\beta_{1} = 0.9$, $\beta_{2} = 0.98$ and $\varepsilon = 10^{-9}$. The learning rate schedule is consistent with~\citet{vaswani2017attention}.

\paragraph{Evaluation}
We measure our translation quality by tokenized case-senstive BLEU~\citep{papineni2002bleu} with \texttt{multi-bleu.pl}\footnote{https://github.com/moses-smt/mosesdecoder/blob/ master/scripts/generic/multi-bleu.perl} for De$\leftrightarrow$En and \texttt{sacreBLEU}\footnote{https://github.com/awslabs/sockeye/tree/master/\\ contrib/sacrebleu} for Ro$\leftrightarrow$En, which is consistent with previous methods. During inference, we generate target tokens autoregressively and use beam search with $beam=6$ and length penalty $\alpha=1.1$. Larger BLEU score indicate better translation quality.

\paragraph{Model Adaption}
We investigate the task difficulty by comparing the training effort of the encoder and decoder from two perspectives in the paper: 1) We vary the number of layers for the encoder and decoder respectively. All models are trained with the same configuration as described before except for the number of layers. 2) For training the encoder side, we initialize the decoder side with the parameters of a well-trained decoder, and initialize other components with random variable. For training decoder side, we follow the opposite operation.

\subsubsection{Experimental Settings for Section 4}

In Section 4, we do not train models from scratch and only evaluate well-trained models with different kind of perturbation to explore the robustness of the encoder and decoder. 

\paragraph{Datasets}
For IWSLT14 De$\leftrightarrow$En and IWSLT14 Ro$\leftrightarrow$En, we use the same validation and test sets as before. For WMT17 Chinese$\leftrightarrow$English (Zh$\leftrightarrow$En)\footnote{http://www.statmt.org/wmt17/translation-task.html}, there are $24M$ sentence pairs for training, $2K$ for both validation and test set. Sentences are encoded using sub-word types based on Byte-Pair-Encoding (BPE)~\citep{sennrich2016neural}\footnote{https://github.com/rsennirich/subword-nmt}, which has $40K$ BPE tokens for the source vocabulary and $37K$ BPE tokens for the target vocabulary. We use newsdev2017 and newstest2017 as validation and test sets respectively.

\paragraph{Model Configuration}
For IWSLT14 De$\leftrightarrow$En and IWSLT14 Ro$\leftrightarrow$En, we use the same model configurations as before. The number of layers for the encoder and decoder are both fixed to $6$. For WMT17 Zh$\leftrightarrow$En translation,  we use \textit{transformer\_big} setting. The model hidden size, feed-forward hidden size and the number of head are set to $1024$, $4096$ and $16$ respectively. We train the model for Zh$\leftrightarrow$En on $8$ NVIDIA TESLA P40 GPUs until convergence. The optimizer and learning rate schedule is consistent with IWSLT14 transaltion tasks.

\paragraph{Evaluation}
For IWSLT14 De$\leftrightarrow$En and IWSLT14 Ro$\leftrightarrow$En, we use the same evaluation metrics as before. For WMT17 Zh$\leftrightarrow$En translation, we calculate the detokenized BLEU score by \texttt{sacreBLEU}. As described in the paper, to eliminate the influence of error propagation, we apply teacher forcing~\citep{williams1989learning} in the inference phase by feeding the previous ground-truth target tokens instead of the previously generated tokens. Therefore, we just add the noise to the ground-truth target tokens when adding noise to the decoder input. We use greedy inference for all settings and length penalty $\alpha=1.1$.

\paragraph{Model Adaption}
For consistency, we use the same well-trained model for each translation task and simulate the input noise with three typical operations to explore the characteristics for the encoder and decoder. In Table~\ref{tab:bleu_score}, we demonstrate the BLEU scores of all the models we used in Section 4 of the paper.

\begin{table}[h]
	\centering
	\caption{The BLEU scores of the models we used on various translation tasks. All models are evaluated with greedy inference and $\alpha=1.1$. TF indicates the model equipped with teacher forcing. (N) indicates the results from non-autoregressive NMT model.}
	\begin{tabular}{c|ccccc}
		\toprule
		&   Transformer & TF \\
		\hline
		De$\to$En &   $32.98$ & $34.79$ \\
		En$\to$De   & $27.25$ & $29.95$ \\
		Ro$\to$En   & $28.84$ & $29.98$ \\
		En$\to$Ro   & $23.10$ & $26.52$ \\
		Zh$\to$En   & $23.20$ & $29.10$ \\
		En$\to$Zh   & $35.10$ & $38.50$ \\
		\hline
		De$\to$En (N) &   $24.06$ & - \\
		En$\to$De (N) &   $16.86$ & - \\
		\bottomrule
	\end{tabular}
	\label{tab:bleu_score}
\end{table}

\subsubsection{Experimental Settings for Section 5}

We give an explanation on why decoder is easier and sensitive in Section 5 of the paper. We additionally train a non-autoregressive NMT model for comparison.

\paragraph{Datasets}
We use the same validation and test sets as before for IWSLT14 translation tasks.

\paragraph{Model Configuration}
We use the same model configurations as before for autoregressive model. As to non-autoregressive counterpart, we use the same network architecture as in NART~\citep{gu2017non}. Both autoregressive and non-autoregressive model are based on \textit{transformer\_small} setting and the number of layers for the encoder and decoder are fixed to $6$. The model hidden size, feed-forward hidden size and the number of head are set to $256$, $1024$ and $4$ respectively. We train non-autoregressive NMT model on one NVIDIA GTX1080Ti GPU until convergence. The optimizer and learning rate schedule is the same as before.

\paragraph{Evaluation}
We use the same evaluation metrics for autoregressive model and non-autoregressive model. We evaluate both models with greedy inference and $\alpha=1.1$. The BLEU scores are demonstrated in Table~\ref{tab:bleu_score}.

\paragraph{Model Adaption}
In section 5 of the paper, we study the importance of previous tokens by dropping the input tokens $w_{t-n}$ for the decoder, where $w_t$ is the current token in target sentence and $n \in [1,t]$ is the distance to the token dropped. We drop the same number of tokens for each $n$. We also compare the autoregressive model with non-autoregressive counterpart by randomly dropping the tokens from the encoder input, encoder output and decoder input. The dropping operation is consistent with Section 4 of the paper.

\subsection{How POS Influences the Encoder and Decoder}

Different kinds of Part-Of-Speech (POS) play different roles in a sentence. For example, nouns typically represents objects and verbs typically conveys actions. To study the impact of POS on the encoder and decoder in NMT, we drop the tokens belongs to each kind of POS and observe the accuracy changes on different kinds of POS.

Due to the definition of POS varies among different languages, we focus on our analysis on English for simplicity and evaluate our trained model on both De$\to$En and En$\to$De for fair comparison, i.e., we drop the input tokens of the encoder for En$\to$De translation, and drop the input tokens of the decoder for De$\to$En translation.

\begin{figure}[t]
	\centering
	\begin{subfigure}[t]{\columnwidth}
		\centering
		\includegraphics[width=0.95\linewidth]{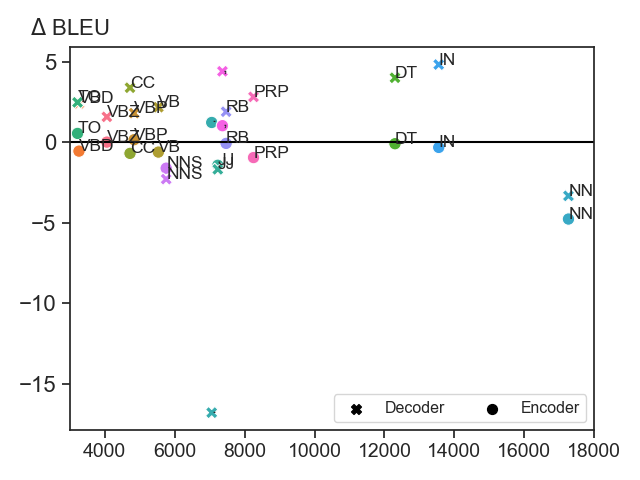}
		\caption{The overall results on main POS types.}
	\end{subfigure}
	\hfill
	\begin{subfigure}[t]{\columnwidth}
		\centering
		\includegraphics[width=\linewidth]{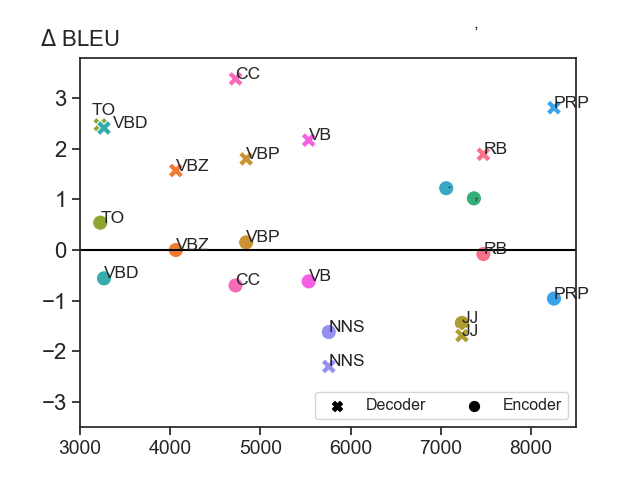}
		\caption{Detailed results.}
	\end{subfigure}
	\caption{The sensitivity of different kinds of POS for the encoder and the decoder. X-axis represents the number of dropped tokens. Y-axis represents the delta BLEU between dropping different kinds of POS and the null baseline. Negative $\Delta$ BLEU score represents that dropping a certain type of POS results in worse accuracy than the null baseline, which indicates the encoder or the decoder is more sensitive to this kind of POS, and vice versa. }
	\label{fig:anal_pos}
\end{figure}

We show the results of the popular POS types (the POS that covers more than $3000$ tokens in the test set) in Figure~\ref{fig:anal_pos}. Since different POS cover different number of tokens in the test set, the direct comparison between different POS is unfair. We introduce a null baseline by randomly dropping different number of tokens and observe the accuracy changes. Based on this data, we fit the curve between the translation accuracy and the number of dropping tokens. We compare each POS with the null baseline by computing the delta BLEU between the accuracy when dropping this kind of POS and the accuracy when dropping tokens randomly. If the delta BLEU is less than 0 (dropping a certain POS achieves lower BLEU score than the null baseline), the model is supposed to be more sensitive to this kind of POS. 

The results are shown in Figure~\ref{fig:anal_pos}. It can be seen that both the encoder and decoder are more sensitive to noun (NN) and adjective (JJ). However, compared with the encoder, the decoder is less sensitive to preposition (IN), determiner (DT), pronoun (PRP), adverb (RB), verb (VB) and coordinating conjunction (CC) but much more sensitive to EOS, which is used to determine the end of a sentence.

\end{document}